\documentclass[lettersize,journal]{IEEEtran}
\usepackage{amsmath,amsfonts}
\usepackage{algorithmic}
\usepackage{algorithm}
\usepackage{array}
\usepackage[caption=false,font=normalsize,labelfont=rm,textfont=rm]{subfig}
\usepackage{textcomp}
\usepackage{stfloats}
\usepackage{url}
\usepackage{verbatim}
\usepackage{graphicx}
\usepackage{cite}
\usepackage{amsmath}
\usepackage[table]{xcolor}
\usepackage{ragged2e} 
\usepackage{booktabs,makecell, multirow, tabularx}
\usepackage{CJKutf8}
\usepackage{booktabs}
\usepackage{multirow}
\usepackage{makecell}
\usepackage{stfloats}
\usepackage{amsmath}
\usepackage{xcolor} 
\usepackage{hyperref} 

\makeatletter
\newcommand{\coloredcite}[1]{{\color{blue}\cite{#1}}}
\makeatother

\hypersetup{
    colorlinks=True,    
    linkcolor=blue,      
    citecolor=blue,    
    filecolor=magenta,  
    urlcolor=magenta,       
    citebordercolor=violet,
  filebordercolor=red,
  linkbordercolor=blue
}

\begin{document}

\title{Life-IQA: Boosting Blind Image Quality Assessment through GCN-enhanced Layer Interaction and MoE-based Feature Decoupling}

\author{Long Tang, Guoquan Zhen,Jie Hao,Jianbo Zhang,Huiyu Duan, Liang Yuan,Guangtao Zhai
\thanks{

Long Tang, Jianbo Zhang, Huiyu Duan are with the College of Integrated Circuits, Shanghai Jiao Tong University, Shanghai 200240, China (e-mail:sjtu8126288@sjtu.deu.cn)

Guoquan Zhen, Jie Hao are with the College  of Integrated Circuits, Beijing University of Chemical Technology, Beijing 100029, China.

Liang Yuan and Guangtao Zhai are with the ICCI and the Institute
of Image Communication and Information Processing, Shanghai Jiao Tong
University, Shanghai 200240, China (e-mail: lyuan@sjtu.edu.cn; zhaiguang-
tao@sjtu.edu.cn).
}
}



\maketitle

\begin{abstract}
Blind image quality assessment (BIQA) plays a crucial role in evaluating and optimizing visual experience. Most existing BIQA approaches fuse shallow and deep features extracted from backbone networks, while overlooking the unequal contributions to quality prediction. Moreover, while various vision encoder backbones are widely adopted in BIQA, the effective quality decoding architectures remain underexplored. To address these limitations, this paper investigates the contributions of shallow and deep features to BIQA, and proposes a effective quality feature decoding framework via GCN-enhanced \underline{l}ayer\underline{i}nteraction and MoE-based \underline{f}eature d\underline{e}coupling, termed \textbf{(Life-IQA)}. Specifically, the GCN-enhanced layer interaction module utilizes the GCN-enhanced deepest-layer features as query and the penultimate-layer features as key, value, then performs cross-attention to achieve feature interaction. Moreover, a MoE-based feature decoupling module is proposed to decouple fused representations though different experts specialized for specific distortion types or quality dimensions. Extensive experiments demonstrate that Life-IQA shows more favorable balance between accuracy and cost than a vanilla Transformer decoder and achieves state-of-the-art performance on multiple BIQA benchmarks.The code is available at: 
\href{https://github.com/TANGLONG2/Life-IQA/tree/main}{\texttt{Life-IQA}}.

\end{abstract}


\section{Introduction}
\label{sec:intro}
\begin{figure}[t]
\label{sec:intro}
\centering
\includegraphics[height=5cm,width=8cm]{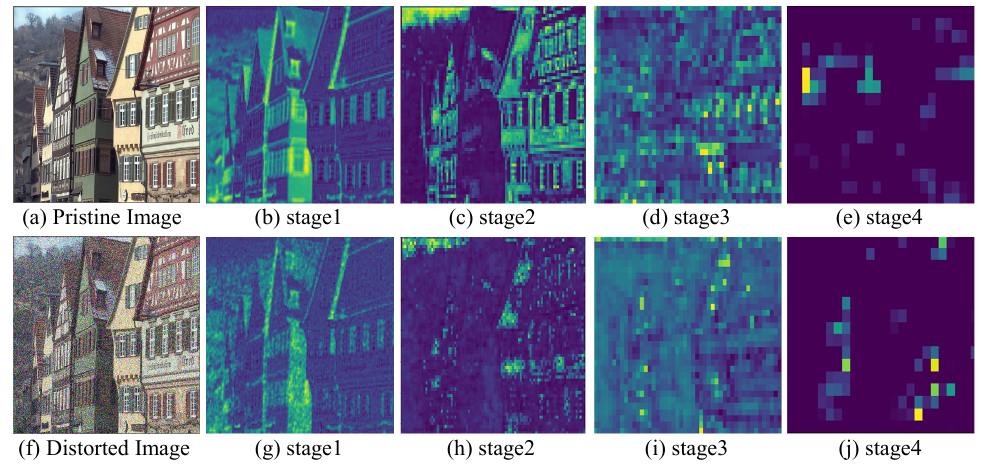}
\caption{Distortion feature map comparison at different stages for a pristine distorted image and a pristine image. Features are more discriminative at the later stages.}
\vspace{-3mm}
\label{fig_1}
\end{figure}

\begin{figure}[t]
\label{sec:intro}
\centering
\includegraphics[height=5cm]{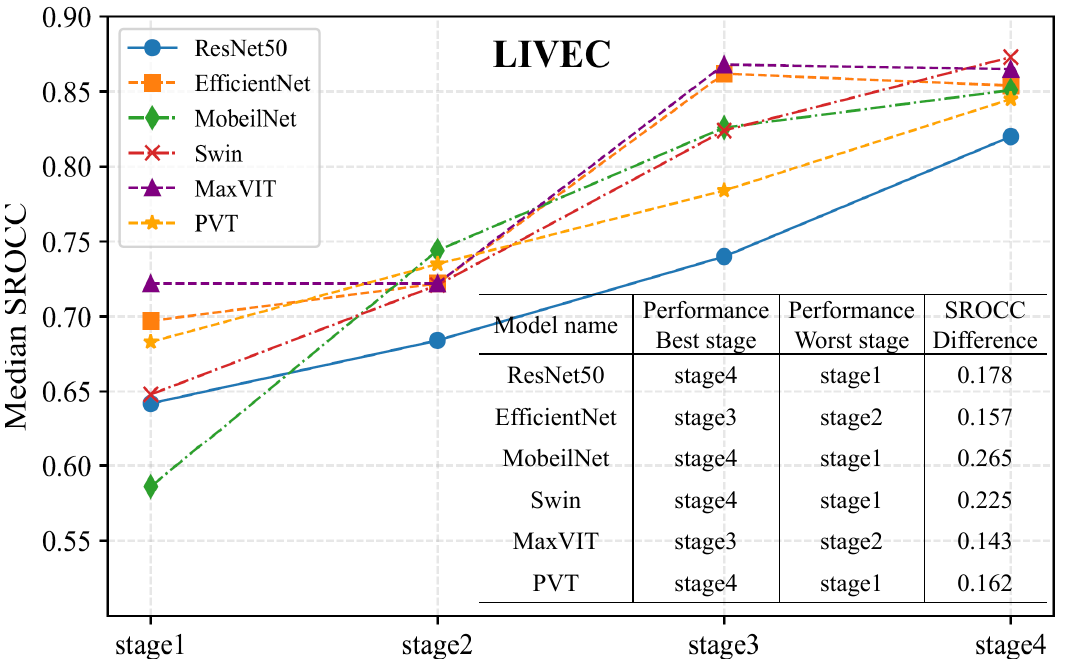}
\caption{Performance of each stage features from Different pretrained models on the LIVEC. Later stages (stage3 and stage4) show better contributions.}
\vspace{-5mm}
\label{fig_2}
\end{figure}
\begin{figure*}[!t]
\centering
\includegraphics[height=7cm,width=18cm]{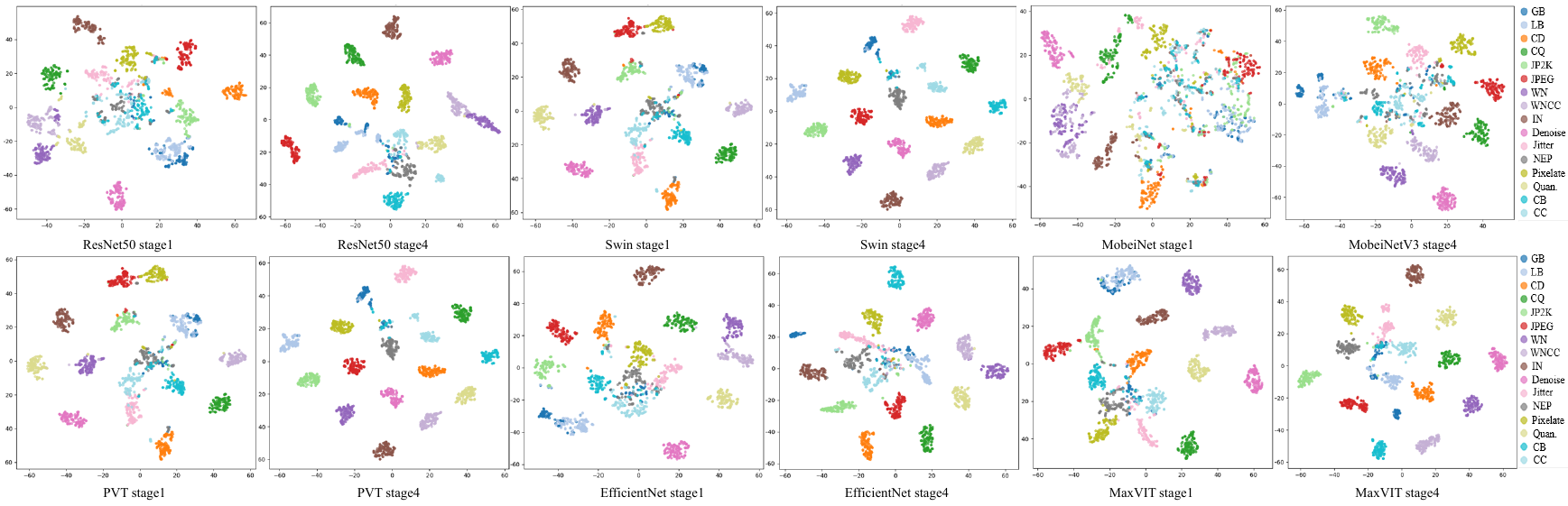}
\caption{t-SNE Visualization of Stage 1 and Stage 4 Features from Different Pretrained Models on the KADID-10k Dataset.}
\vspace{-5mm}
\label{fig_3}
\end{figure*}

Image quality assessment (IQA) has long been an important task in computer vision, which aims to accurately produce quality predictions that are consistent with human perceptual assessments \coloredcite{10109108,11092710,11093358,11175315,10795128,10056825,9961939}. Among these, blind image quality assessment (BIQA) qualifies the absolute quality of images in the absence of pristine references \coloredcite{10960664,10647859,10833846,10463077,10462930}, which is widely used in various applications. Existing BIQA methods commonly utilize pretrained backbone networks to extract multi-scale features from distorted images then fuse them to predict the quality \coloredcite{9156687,9817377,9121773,10323206,10036133}. The underlying assumption of this strategy is that different hierarchical features provide complementary information. Specifically, shallow level features are thought to capture local details and texture information, making them sensitive to local distortions, as exemplified by $stage1\in \mathbb{R}^{\frac{H}{4}\times \frac{W}{4}\times C_1}$ and $stage2\in \mathbb{R}^{\frac{H}{8}\times \frac{W}{8}\times C_2}$ in Figure \ref{fig_1}, where $H,W$ respectively represent the height and width of the original image and $C_i$ denotes the channel size. In contrast, deep level features encode global and semantic information, making them more effective at reflecting content-related distortions, as exemplified by $stage3\in \mathbb{R}^{\frac{H}{16}\times \frac{W}{16}\times C_3}$ and $stage4\in \mathbb{R}^{\frac{H}{32}\times \frac{W}{32}\times C_4}$ in Figure \ref{fig_1}. However, the contributions of different layers to quality prediction remain underexplored. 

To quantify the actual contribution of features from different stages to the BIQA, six different pretrained models are evaluated on the authentic distortion dataset LIVEC \coloredcite{8478807}, including three convolutional neural
networks (CNNs), \textit{i.e.}, ResNet50 \coloredcite{he2015deepresiduallearningimage}, EfficientNet \coloredcite{tan2020efficientnetrethinkingmodelscaling}, MobileNet \coloredcite{howard2019searchingmobilenetv3} and three Transformer models, \textit{i.e.}, Swin Transformer \coloredcite{liu2021swintransformerhierarchicalvision}, MaxViT \coloredcite{tu2022maxvitmultiaxisvisiontransformer}, PVT \coloredcite{wang2021pyramidvisiontransformerversatile}. To separately investigate the quality-representation capacity of the features from different layers, the image features from each stage are directly appended with a simple prediction head, which consists of a global average pooling layer (GAP) and a linear regression layer. As illustrated in Figure \ref{fig_2}, the regression results on the LIVEC \coloredcite{8478807} clearly demonstrate that the predictive performance of shallow level features ($stage1, stage2$) from all backbone networks is significantly inferior to that of deep-level features ($stage3, stage4$). To further investigate the discriminative ability of the features for specific distortion types, the features from $stage1$ and $stage4$ are visualized using t-SNE on the synthetic distortion dataset KADID-10k \coloredcite{8743252}. As shown in Figure \ref{fig_3}, the features from $stage1$ are entangled and hard to be completely distinguished, leading to severely blurred class boundaries, while the $stage4$ features form highly compact and well-separated cluster structures. These experiments indicate that shallow features contribute less to BIQA tasks compared with deep features. Because of the limited data in IQA, models struggle to learn effective quality representations from shallow features dominated by details, which indicates that completely and directly fuse features from all stages may introduce noise and redundancy. 




\begin{figure*}[!t]
\label{sec:intro}
\centering
\includegraphics[width=6in]{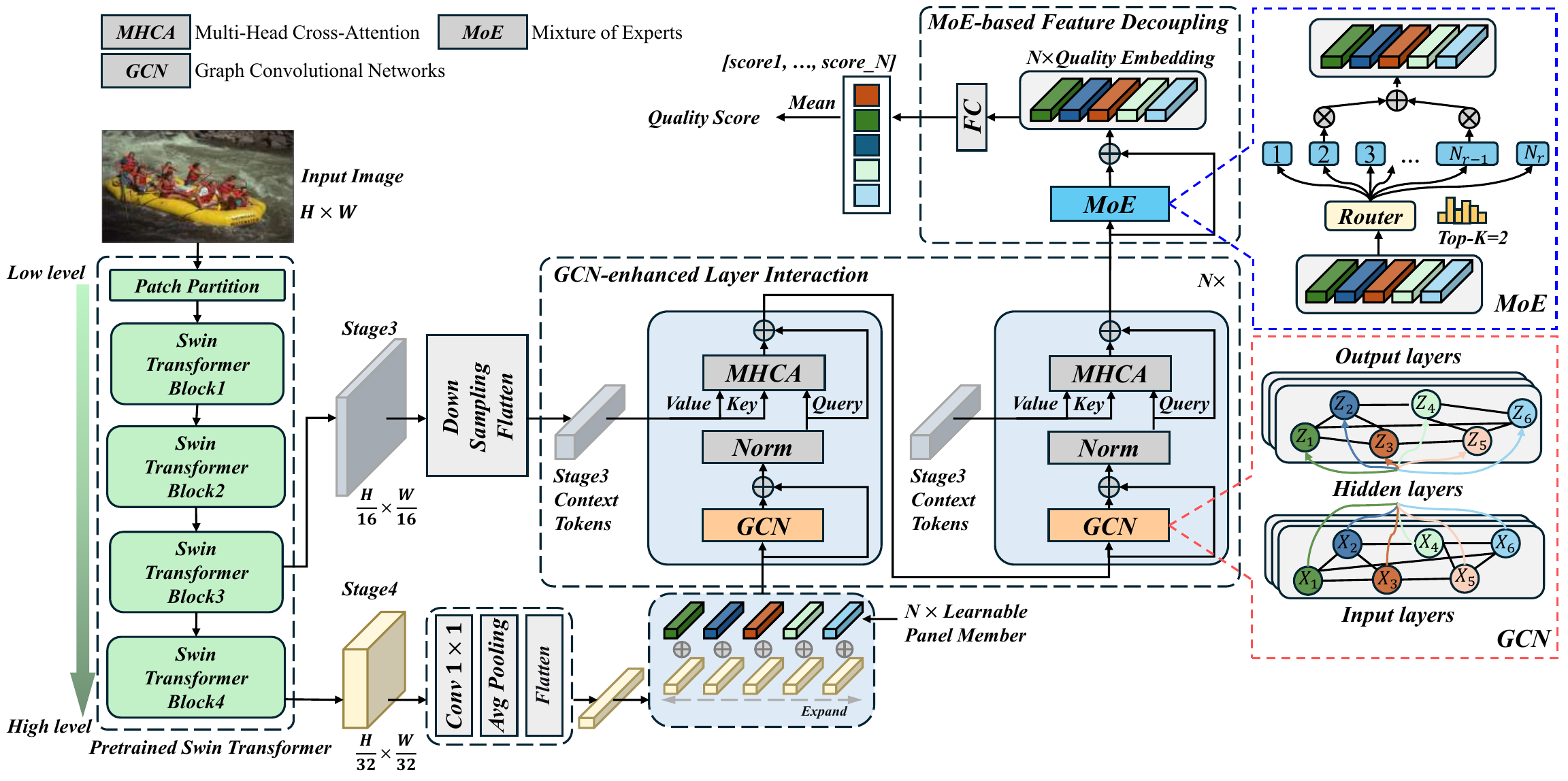}
\caption{Framework overview of proposed Life-IQA. The image is passed through a pretrained Swin Transformer to extract $stage4$ and $stage3$ features. A layer interaction enhanced by a GCN fuses them; a MoE performs decoupling of the fused representation into distortion specific factors; and linear projections for each token are averaged to yield the final quality score.}
\vspace{-5mm}
\label{fig_4}
\end{figure*}

Based on the aforementioned analysis, we avoid the unstable and often ineffective traditional multi-scale feature fusion and focus on enabling efficient interaction among deeper, semantically richer features. To this end, we propose \textbf{Life-IQA}, an effective quality feature decoding framework built upon \textbf{GCN-enhanced Layer Interaction} and \textbf{MoE-based Feature Decoupling}. In Life-IQA, the deepest-stage features, enhanced through a graph convolutional network (GCN), serve as query tokens that carry high-level quality semantics, while the penultimate-stage features act as keys and values, providing structured local evidence, then the cross-attention-based interaction is performed. Furthermore, a mixture-of-expert (MoE) module decouples the fused representations, allowing different experts to specialize in particular distortion types or quality dimensions. This design yields a dynamic, data-efficient decoding mechanism tailored to the characteristics of BIQA. Overall, the main contributions are summarized as follows.

\begin{itemize}
\item We propose a cross-attention based layer interaction method for BIQA, which extracts the features of last two stages and fuse them via cross-attention layers.

\item A lightweight GCN is applied to explicitly capture the topology and dependencies among query tokens of the cross-attention.

\item To handle diverse image distortion types, we introduce a MoE-based feature decoupling module with a set of parallel expert networks, which enables collaborative discrimination of image features across multiple distortion dimensions, improving modeling accuracy for complex mixed distortions and enhancing the robustness of quality assessment. To our knowledge, this is first work that utilizes MoE in BIQA.

\item 
A systematic validation is conducted on seven benchmark datasets, covering a wide range of image content, distortion types, and data scales. The experimental results demonstrate Life-IQA significantly outperforms current mainstream approaches on multiple test sets.
\end{itemize}

\section{Related Work}
\subsection{Learning based Image Quality Assessment}
With the success of the Vision Transformer (ViT) \coloredcite{dosovitskiy2021imageworth16x16words}, the powerful capability for modeling global dependencies has also been applied to the BIQA task \coloredcite{10478301,9506075}. Golestaneh et al \coloredcite{9706735} combined the local feature extraction ability of CNNs with Transformers and introduced a relative ranking loss to enhance the model's perceptual consistency. To address the limitation of Transformers requiring fixed size inputs, Ke et al \coloredcite{9710973} enabled the processing of images with different aspect ratios through techniques such as hash encoding, demonstrating strong performance on large scale datasets. However, the vast majority of current Transformer based BIQA methods focus on the encoder structure for feature extraction and fusion. The potential of the decoder structure for target driven information aggregation via queries remains largely unexplored. DEIQT \coloredcite{qin2023dataefficientimagequalityassessment} marked the first significant attempt in this direction by constructing a complete encoder decoder pipeline. However, the query generation relied solely on the single CLS token output from the ViT encoder. This approach not only limited the diversity of query perspectives but also made the effectiveness of the decoding process heavily dependent on the representation quality of the upstream encoder. Therefore, designing decoders with richer information sources and more flexible structures is a key to unlocking the full potential in BIQA.

\subsection{Mixture of Experts}
The MoE model was a neural network architecture that expands model capacity through conditional computation \coloredcite{11226953,zhou2023brainformers,10.5555/3618408.3618649}. The core idea was a task could be decomposed and handled collaboratively by a set of specialized subnetworks, known as ``experts". A trainable gating network, or router, dynamically selected one or more of the most relevant experts to activate based on the input. This concept was first proposed by Jacobs et al \coloredcite{6797059}, with the goal of allowing different expert networks to specialize in distinct regions of the input space. In recent years, with the explosive growth in model parameter scale, MoE has garnered significant attention as a key technology for overcoming the bottlenecks of dense models. Particularly within the Transformer architecture, replacing the feed-forward network (FFN) with MoE become a mainstream paradigm \coloredcite{lepikhin2020gshardscalinggiantmodels}. More recently, the Mixtral $8\times7B$ model from Mixtral AI \coloredcite{jiang2024mixtralexperts} employed a Top-2 routing strategy, selecting two out of eight experts for each token and further validating the efficacy of sparse activation. Despite the success of MoE in Natural Language Processing \coloredcite{11230232,10377558,liu2025netMoE}, the application of MoE in the domain of BIQA remains an unexplored area. 




\section{The Proposed Method}
\subsection{Overall Architecture}
The overall framework of our method is illustrated in Figure \ref{fig_4} and consists of two core innovative components. The first is a GCN-enhanced
layer interaction, responsible for selective cross level feature fusion. This layer interaction generates quality queries from the deepest features and aggregates key information from the penultimate deep features through cross-attention mechanisms. The second component is a MoE, which decouple the fused features and enables multiple expert networks to collaboratively perceive the final image quality.

\subsection{GCN-enhanced Layer Interaction}
A standard Transformer decoder typically employs a set of learnable embeddings as initial queries $Q_{init}\in \mathbb{R}^{N\times D}$, where $N$ is the number of query tokens; $D$ is the channel size 384. Lacking a favorable starting anchor in the feature space relevant to image content, $Q_{init}$ leads to inefficient learning. To address this bottleneck, $Q_{init}$ embeds the information from $Stage4$, which serves as a compact representation of the global image content. $Stage4$ is first passed through a $1\times1$ convolutional layer ($conv1$) to reduce channel dimensionality to $D$. Subsequent compression via a GAP operation yields a global context vector $v_{global}\in \mathbb{R}^{D}$.  Finally, $v_{global}$ is broadcast to match the length $N$ of the query sequence and added element-wise to $Q_{init}$ to yield the guided initial queries $Q_{init}^{'}\in \mathbb{R}^{N\times {D}}$. This process can be formulated as:
\begin{equation}
Q_{init}^{'}\ = expand\left( GAP\left( conv1\left( stage4 \right) \right) \right) \ +\ Q_{init}
  \label{eq:important}
\end{equation}
where $Q_{init}^{'}$ is the enhanced query sequence. In this way, each query is instilled with preliminary information about the image's overall quality attributes, which significantly accelerates the model's convergence process.

Transformer decoder comprises three residual connection modules: a self-attention mechanism, a cross-attention mechanism, and a FFN. The self-attention mechanism can be formalized by the following equation:
\begin{equation}
Attention\left( Z_{i+1} \right) \ =\ soft\max \left( \frac{Q_{i+1}K_{i+1}^{T}}{\sqrt{d_k}} \right) V_{i+1}  
\label{eq:important}
\end{equation}
where the $Q_{i+1}$, $K_{i+1}$, and $V_{i+1}$ are obtained from the output of the previous layer, $Z_i$ through independent linear projections, i.e.:
\begin{equation}
Q_{i+1}=f_1\left( Z_i \right) ,\ K_{i+1}=f_2\left( Z_i \right) ,\ V_{i+1}=f_3\left( Z_i \right)
\label{eq:important}
\end{equation}
$f_i$ is a linear transformation. $Q_{i+1}$ and $K_{i+1}$ vectors are generated by independently applying these linear transformations to each token in the input sequence. The generation process is token-wise and non-interactive, so each token during the projection stage is unaware of the existence of other tokens. This approach leads to a lack of collaborative nature and informational complementarity in the queries, limiting the model's ability to capture complex, structured dependencies within the queries themselves.

To solve this problem, a GCN is designed to replace the query and key interaction process in conventional self-attention. The core idea of this module is to establish an explicit information propagation path among the query vectors. Specifically, the $Q_{init}^{'}\in \mathbb{R}^{N\times {D}}$ is represented as $N$ nodes in a graph, where each node has a feature dimension of $D$. The graph's structure is defined by an adjacency matrix $A_{i}\in \mathbb{R}^{N\times N}$. $A_i$ is a learnable matrix, allowed to adaptively capture the intrinsic relationships between queries. The propagation rule for each layer can be formulated as:
\begin{align}
\mathcal{Q}_1 &= f_{input}(Q_{init}^{'}, A_{1}) = \sigma(A_{1} \cdot Q_{init}^{'} \cdot W_0) \\
\mathcal{Q}_2 &= f_{hidden}(H_1, A_{2}) = \sigma(A_{2} \cdot H_1 \cdot W_1) \\
\mathcal{Q}_3 &= f_{output}(H_2, A_{3}) = A_{3} \cdot H_2 \cdot W_2
\end{align}
where $H_{i}$  is the node representation at layer $i$, $W_{i}\in \mathbb{R}^{^{D\times D}}$ is the trainable weight matrix for this layer, and $\sigma $  is a non-linear activation function. After processing through three layers of the GCN,  $\mathcal{Q}_3$ incorporates neighborhood information, possesses a richer contextual awareness, and will be used for the subsequent cross-attention computation.

Benefiting from the asymmetry of cross-attention, query and key/value need not be aligned in scale during fusion. Therefore, the features from $stage3$ are used as the key/value to provide supplementary information for $\mathcal{Q}_3$. Since flattening $stage3$ into a sequence introduces substantial computational overhead, this paper adopts the partition average pooling strategy \coloredcite{10222634}. $Stage3$ is divided into an $N\times N$ grid, where average pooling is performed within each grid region. This operation efficiently transforms the high dimensional feature map into a compact feature sequence $S\in \mathbb{R}^{N^2\times D}$, where each element represents the average feature response of a local region in the original image. The cross-attention module takes $\mathcal{Q}_3$ and $S$ as inputs. Within this module, $\mathcal{Q}_3$ serves as the Query, while the sequence $S$, acts as the information source from which the key and value are generated. Through the attention computation, each query vector adaptively aggregates the most relevant semantic information from the sequence $S$, thereby achieving effective cross-level integration of complementary deep features.

\begin{figure}[!t]
\centering
\includegraphics[width=3.3in]{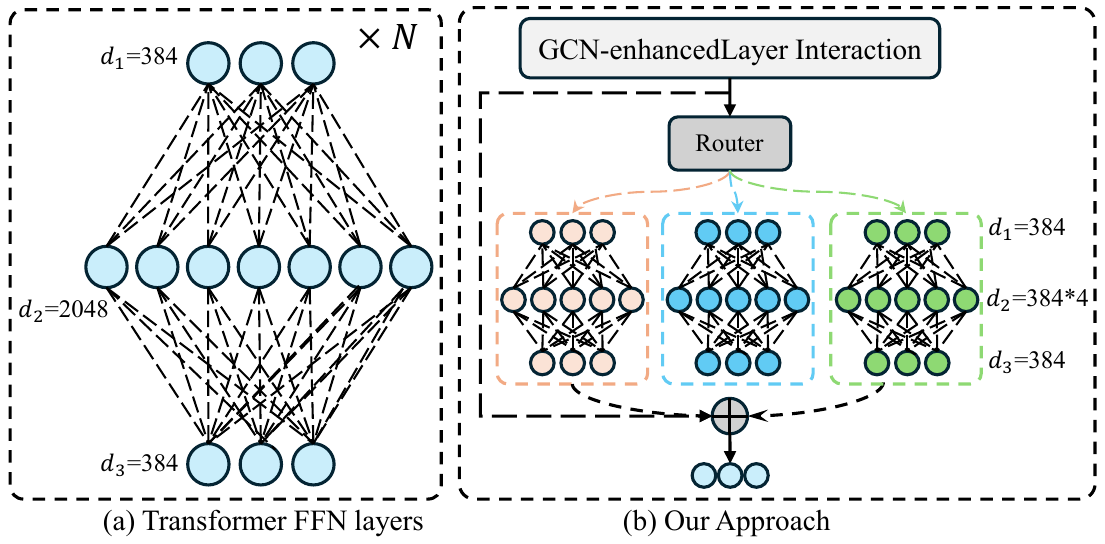}
\caption{Comparison between standard FFN and the proposed MoE-based Feature Decoupling.}
\vspace{-5mm}
\label{fig_5}
\end{figure}
\subsection{MoE-based Feature Decoupling}
As illustrated in Figure \ref{fig_5}, the FFN in a standard Transformer decoder consists of two linear transformations with both input and output dimensions of 384 and a hidden dimension of 2048, resulting in approximately $1.58 M$ parameters.The FFN accounts for approximately 70\% of the total parameters in the Transformer decoder. The fixed and input-independent mapping mechanism of FFN limits representational flexibility and increases the risk of overfitting on small scale IQA datasets. Although most existing methods embed MoE directly into the decoder layer as an FFN equivalent module, the overall parameter count still grows linearly with the number of experts, even when the hidden size of each expert is reduced to 1536 or 768. Specifically, Configurations with 2, 4, and 8 experts contain approximately $2.36M$, $4.73M$, and $9.46M$ parameters, which are all considerably larger than those of the standard FFN. Consequently, without large scale pretraining, MoE modules fail to effectively optimize their sparse activation mechanisms on limited IQA datasets, often resulting in performance comparable to that of the original FFN. Therefore, in the Life-IQA, the MoE head is positioned at the end of the decoding process specifically to make a final ensemble decision on the highly refined features. This ``post-posed" design enables the MoE to achieve more robust quality judgments without relying on expert pretraining. 


As shown in Figure \ref{fig_4}, the MoE block consists of a gating network $G$ and expert networks $\left\{ \mathcal{E}_e \right\} _{e=1}^{N_E}$ and a routing mechanism, where each expert $ \mathcal{E}_e $ is an independent Multi-Layer Perceptron. Given inputs $X\in \mathbb{R}^{N\times D}$, the gate estimates token–expert affinities and ranks experts per token. To realize sparse activation, only the $Top$-$K$ candidates are kept, with non-$Top$-$K$ logits masked to $-\infty$, followed by a softmax operation to produce sparse routing weights. Each token has only $K$ non-zero weights over experts, which are used to aggregate the selected expert outputs:
\begin{align}
g_i&=G(X_i) = W_g \cdot X_{i} + b_g,i\in \left[ 1,N \right]  \\
\mathcal{M}_{i,e}& = \begin{cases} 
1 & \text{if } e \in \text{TopK}(g_i,K) \\
0 & \text{otherwise}
\end{cases}\\
\tilde{g}_{i,e} &=\left\{ \begin{array}{l}
	g_i \ \ if\ \mathcal{M}_{i,e}=1\\
	-\infty \ otherwise\\
\end{array} \right. \\
Y_{i}\ &=\sum\nolimits_{e=1}^{E}{soft\max_e \left( \tilde{g}_{i}\right)}\cdot \mathcal{E}_e\left( X_i \right) 
\label{eq:gating}
\end{align}
where $W_g\in \mathbb{R}^{D\times E}$ is the trainable gating weight matrix. $Y_{i}$ is the weighted sum of the i token outputs from all experts, $\underset{e}{soft\max}$ is applying softmax along the expert dimension $e$. To prevent the scores of certain experts from becoming overly concentrated during training, a router $z$-loss $\mathcal{L}_z$ and a load-balancing loss $\mathcal{L}_{aux}$ are incorporated into the training process of Life-IQA to balance the workload among experts. The corresponding loss functions are formulated as follows:
\vspace{-4mm}
\begin{align}
p_{i,e}=\underset{e}{soft\max}\left( g_i \right) \ ,\ \hat{p}_e=\frac{1}{N}\sum\nolimits_{i=1}^N{p_{i,e}}\\
\hat{t}_e=\frac{1}{N}\sum\nolimits_{i=1}^N{\mathcal{M}_{i,e}}\ ,\ \mathcal{L}_{aux}={N_E}\cdot \sum\nolimits_{e=1}^{N_E}{\hat{t}_e\hat{p}_e}\\
\mathcal{L}_z=\frac{1}{N}\sum\nolimits_{i=1}^N{\left(\log \sum\nolimits_{e=1}^{N_E}{\exp \left( g_i \right)}\right)^2}
\label{eq:important}
\end{align}




To further enhance training stability and suppress the gradient variance that may arise from routing decisions, a regularization bypass branch is employed. This branch, controlled by a learnable scalar coefficient $\gamma $, creates a residual connection between the MoE output and the original input. The final output $Y_{final}$ is calculated as follows:
\begin{equation}
Y_{final}=Y_{MoE}\left( X \right) +\gamma \cdot X
\label{eq:important}
\end{equation}
This output sequence, $Y_{final}$, is then passed through a linear layer and a GAP to produce the final image quality score. 

\subsection{Mixture of Experts Head}
As illustrated in Figure \ref{fig_5}, the FFN in a standard Transformer decoder consists of two linear transformations with both input and output dimensions of 384 and a hidden dimension of 2048, resulting in approximately $1.58 M$ parameters.The FFN accounts for approximately 70\% of the total parameters in the Transformer decoder. The fixed and input-independent mapping mechanism of FFN limits representational flexibility and increases the risk of overfitting on small-scale IQA datasets. Although most existing methods embed MoE directly into the decoder layer as an FFN-equivalent module, the overall parameter count still grows linearly with the number of experts, even when the hidden size of each expert is reduced to 1536 or 768. Specifically, Configurations with 2, 4, and 8 experts contain approximately $2.36M$, $4.73M$, and $9.46M$ parameters, which are all considerably larger than those of the standard FFN. Consequently, without large-scale pre-training, MoE modules fail to effectively optimize their sparse activation mechanisms on limited IQA datasets, often resulting in performance comparable to that of the original FFN. Therefore, in the GTDM, the MoE head is positioned at the end of the decoding process specifically to make a final ensemble decision on the highly refined features. This ``post-posed" design enables the MoE to achieve more robust quality judgments without relying on expert pretraining.

As shown in Figure \ref{fig_4} (b), the MOE block consists of a gating network $G$ and expert networks $\left\{ \mathcal{E}_e \right\} _{e=1}^{N_E}$ and a routing mechanism, where each expert $ \mathcal{E}_e $ is an independent Multi-Layer Perceptron. Given inputs $X\in \mathbb{R}^{N\times D}$, the gate estimates token–expert affinities and ranks experts per token. To realize sparse activation, only the $Top$-$K$ candidates are kept, with non-$Top$-$K$ logits masked to $-\infty$, followed by a softmax operation to produce sparse routing weights. Each token has only $K$ non-zero weights over experts, which are used to aggregate the selected expert outputs:
\begin{align}
g_i&=G(X_i) = W_g \cdot X_{i} + b_g,i\in \left[ 1,N \right]  \\
\mathcal{M}_{i,e}& = \begin{cases} 
1 & \text{if } e \in \text{TopK}(g_i,K) \\
0 & \text{otherwise}
\end{cases}\\
\tilde{g}_{i,e} &=\left\{ \begin{array}{l}
	g_i \ \ if\ \mathcal{M}_{i,e}=1\\
	-\infty \ otherwise\\
\end{array} \right. \\
Y_{i}\ &=\sum\nolimits_{e=1}^{E}{soft\max_e \left( \tilde{g}_{i}\right)}\cdot \mathcal{E}_e\left( X_i \right) 
\label{eq:gating}
\end{align}
where $W_g\in \mathbb{R}^{D\times E}$ is the trainable gating weight matrix. $Y_{i}$ is the weighted sum of the i token outputs from all experts, $\underset{e}{soft\max}$ is applying softmax along the expert dimension $e$. To prevent the scores of certain experts from becoming overly concentrated during training, a router $z$-loss $\mathcal{L}_z$ and a load-balancing loss $\mathcal{L}_{aux}$ are incorporated into the training process of GTDM to balance the workload among experts. The corresponding loss functions are formulated as follows:

\begin{align}
p_{i,e}=\underset{e}{soft\max}\left( g_i \right) \ ,\ \hat{p}_e=\frac{1}{N}\sum\nolimits_{i=1}^N{p_{i,e}}\\
\hat{t}_e=\frac{1}{N}\sum\nolimits_{i=1}^N{\mathcal{M}_{i,e}}\ ,\ \mathcal{L}_{aux}={N_E}\cdot \sum\nolimits_{e=1}^{N_E}{\hat{t}_e\hat{p}_e}\\
\mathcal{L}_z=\frac{1}{N}\sum\nolimits_{i=1}^N{\left(\log \sum\nolimits_{e=1}^{N_E}{\exp \left( g_i \right)}\right)^2}
\label{eq:important}
\end{align}




To further enhance training stability and suppress the gradient variance that may arise from routing decisions, a regularization bypass branch is employed. This branch, controlled by a learnable scalar coefficient $\gamma $, creates a residual connection between the MoE output and the original input. The final output $Y_{final}$ is calculated as follows:
\begin{equation}
Y_{final}=Y_{moe}\left( X \right) +\gamma \cdot X
\label{eq:important}
\end{equation}
This output sequence, $Y_{final}$, is then passed through a linear layer and a GAP to produce the final image quality score. 

\subsection{Quality Score Regression and Loss Function}
To map the final feature $Y_{final}$ to the quality score of the image, we feed it into a linear regression layer to obtain the predicted quality score, as illustrated in Figure \ref{fig_4}. Each query token produces an individual quality score, and the final image quality score is obtained by averaging the scores of all tokens.

The total loss function is composed of three main components: the L1 loss $\mathcal{L}_{main}$ for the image quality scores, a load balancing loss $\mathcal{L}_{aux}$, and a routing loss $\mathcal{L}_z$, which are included to balance expert utilization and load. The overall loss function is shown as follows:
\begin{align}
\mathcal{L}_{\text{main}} &= \frac{1}{N} \sum\nolimits_{i=1}^N (\hat{y}_i - y_i) \label{eq:main} \\
\mathcal{L}_{\text{total}} &= \mathcal{L}_{\text{main}} + \lambda_1 \cdot \mathcal{L}_{\text{aux}} + \lambda_2 \cdot \mathcal{L}_z \label{eq:total}
\end{align}
The loss weights are set to $\lambda_{1}=0.01$ and $\lambda_{2}=0.001$.

\begin{table*}[htbp]
  \centering
  \renewcommand{\arraystretch}{1.2}
   \caption{Comparison with multiple advanced IQA methods on several benchmark datasets.
  SROCC and PLCC represent Spearman and Pearson correlation coefficients, respectively.
  The best and second-best results are highlighted in \textbf{\textcolor{red}{red}} and \textbf{\textcolor{blue}{blue}}.}
  \scalebox{0.85}{
  \begin{tabular}{lcccccccccccccccc}
    \toprule[0.5mm]
    {Method} & \multicolumn{1}{c}{Params} & \multicolumn{2}{c}{LIVE} & \multicolumn{2}{c}{CSIQ} & \multicolumn{2}{c}{TID2013} & \multicolumn{2}{c}{KADID-10K} & \multicolumn{2}{c}{LIVEC} & \multicolumn{2}{c}{KonIQ} & \multicolumn{2}{c}{SPAQ} \\
    \cmidrule{3-16}
    \multicolumn{1}{c}{} & \multicolumn{1}{c}{M} &
    \multicolumn{1}{c}{SROCC} & \multicolumn{1}{c}{PLCC} &
    \multicolumn{1}{c}{SROCC} & \multicolumn{1}{c}{PLCC} &
    \multicolumn{1}{c}{SROCC} & \multicolumn{1}{c}{PLCC} &
    \multicolumn{1}{c}{SROCC} & \multicolumn{1}{c}{PLCC} &
    \multicolumn{1}{c}{SROCC} & \multicolumn{1}{c}{PLCC} &
    \multicolumn{1}{c}{SROCC} & \multicolumn{1}{c}{PLCC} &
    \multicolumn{1}{c}{SROCC} & \multicolumn{1}{c}{PLCC} \\
    \midrule
    MUSIQ \coloredcite{9376644} & 27 &
      0.940 & 0.911 & 0.871 & 0.893 & 0.773 & 0.815 & 0.875 & 0.872 & 0.702 & 0.746 & 0.916 & 0.928 & 0.918 & 0.921 \\
    TRes \coloredcite{9706735} & 152 &
      0.969 & 0.968 & 0.922 & 0.942 & 0.863 & 0.883 & 0.859 & 0.858 & 0.846 & 0.877 & 0.915 & 0.928 &  -  &  -  \\
    DACNN \coloredcite{9817377} & - &
      0.978 & 0.980 & 0.943 & 0.957 & 0.871 & 0.889 & 0.905 & 0.905 & 0.866 & 0.884 & 0.901 & 0.912 & 0.915 & 0.921 \\
    DEIQT \coloredcite{qin2023dataefficientimagequalityassessment} & 24 &
      0.980 & \textbf{\textcolor{blue}{0.982}} & 0.946 & 0.963 & 0.892 & 0.908 & 0.889 & 0.887 & 0.875 & 0.894 & 0.921 & 0.934 & 0.919 & 0.923 \\
   CICI \coloredcite{10213444} & - &
      0.977 & 0.979 & 0.930 & 0.951 & 0.865 & 0.874 &  -  &  -  & 0.865 & 0.874 & 0.921 & 0.930 &  -  &  -  \\
    PCIQA \coloredcite{10520932} & - &
      0.976 & 0.976 & 0.931 & 0.942 &  -  &  -  & 0.870 & 0.866 & 0.863 & 0.880 & 0.918 & 0.930 &  -  &  -  \\
    TempQT \coloredcite{10287911} & - &
      0.976 & 0.977 & 0.950 & 0.960 & 0.883 & 0.906 &  -  &  -  & 0.870 & 0.886 & 0.915 & 0.928 &  -  &  -  \\
    FsPN \coloredcite{10706861} & 34 &
      0.979 & 0.979 & 0.955 & 0.962 & 0.897 & 0.912 & 0.911 & 0.909 & 0.860 & 0.884 & 0.918 & 0.932 &  -  &  -  \\
    LIQE \coloredcite{10204315} & 151 &
      0.970 & 0.951 & 0.936 & 0.939 &  -  &  -  & 0.930 & 0.931 & \textbf{\textcolor{red}{0.904}} & 0.910 & 0.919 & 0.908 &  -  &  -  \\
    Re-IQA \coloredcite{10204004} & 48 &
      0.970 & 0.971 & 0.947 & \textbf{\textcolor{blue}{0.960}} & 0.880 & 0.901 & 0.908 & 0.912 &  -  &  -  &  -  &  -  & 0.918 & 0.925 \\
    QCN \coloredcite{10656546} & - &
      -  & -  & -  & -  & -  & -  & -  & -  & 0.875 & 0.893 & 0.934 & 0.945 & 0.923 & 0.928 \\
    LoDa \coloredcite{10655501} & 118 &
      0.975 & 0.979 &  -  & -  & 0.869 & 0.901 & 0.931 & 0.936 & 0.876 & 0.899 & \textbf{\textcolor{blue}{0.932}} & \textbf{\textcolor{blue}{0.944}} & \textbf{\textcolor{blue}{0.925}} & \textbf{\textcolor{blue}{0.928}} \\
    VISGA \coloredcite{10845143} & - &
      \textbf{\textcolor{red}{0.987}} & \textbf{\textcolor{red}{0.988}} &
      \textbf{\textcolor{blue}{0.960}} & \textbf{\textcolor{red}{0.971}} &
      \textbf{\textcolor{blue}{0.901}} & \textbf{\textcolor{blue}{0.914}} &
      \textbf{\textcolor{blue}{0.939}} & \textbf{\textcolor{red}{0.944}} &
      0.882 & \textbf{\textcolor{blue}{0.912}} & 0.931 & 0.940 & - & - \\
    \midrule[0.3mm]
    {Life-IQA} & 95 &
      \textbf{\textcolor{blue}{0.982}} & \textbf{\textcolor{blue}{0.982}} &
      \textbf{\textcolor{red}{0.966}} & \textbf{\textcolor{red}{0.971}} &
      \textbf{\textcolor{red}{0.918}} & \textbf{\textcolor{red}{0.930}} &
      \textbf{\textcolor{red}{0.940}} & \textbf{\textcolor{blue}{0.943}} &
      \textbf{\textcolor{blue}{0.896}} & \textbf{\textcolor{red}{0.919}} &
      \textbf{\textcolor{red}{0.937}} & \textbf{\textcolor{red}{0.948}} &
      \textbf{\textcolor{red}{0.926}} & \textbf{\textcolor{red}{0.929}} \\
    \bottomrule[0.5mm]
  \end{tabular}}
 
  \label{table_1}
  \vspace{-4mm}
\end{table*}



\begin{table}[htbp]
  \centering
   \caption{Cross-dataset evaluation when training on KonIQ.
Performance is evaluated on CSIQ, TID2013, and LIVE.}
  \scalebox{0.9}{
  \renewcommand{\arraystretch}{1.2}
    \begin{tabular}{p{7em}cccccc}
    \toprule[0.4mm]
    \multicolumn{1}{c}{Train on }& \multicolumn{6}{c}{KonIQ} \\
    \midrule
    \multicolumn{1}{c}{Test on} & \multicolumn{2}{c}{CSIQ} & \multicolumn{2}{c}{TID2013} & \multicolumn{2}{c}{LIVE} \\
    \midrule
    \multicolumn{1}{c}{} & \multicolumn{1}{p{2.5em}}{SROCC} & \multicolumn{1}{p{2.5em}}{PLCC} & \multicolumn{1}{p{2.5em}}{SROCC} & \multicolumn{1}{p{2.5em}}{PLCC} & \multicolumn{1}{p{2.5em}}{SROCC} & \multicolumn{1}{p{2.5em}}{PLCC} \\
    \midrule
     DBCNN \coloredcite{8576582} & 0.459 & 0.525 & 0.348 & 0.471 & 0.683 & 0.525 \\
    HyperIQA \coloredcite{9156687}  & 0.505 & 0.574 & 0.396 & 0.508 & 0.622 & 0.679 \\
     DACNN \coloredcite{9817377} & 0.626  & 0.632 & 0.400 & 0.524 & 0.760   & 0.726 \\
    FsPN \coloredcite{10706861}   & 0.711 & 0.734 & 0.403 & 0.495 & \textbf{\textcolor{red}{0.829}} & 0.797 \\
    Life-IQA  &\textbf{\textcolor{red}{0.715}}       & \textbf{\textcolor{red}{0.738}}      &  \textbf{\textcolor{red}{0.453}}     &  \textbf{\textcolor{red}{0.578}}     & 0.822      & \textbf{\textcolor{red}{0.800}} \\
    \bottomrule[0.4mm]
    \end{tabular}%
    }
   
  \label{table_3}%
  \vspace{-6mm}
\end{table}

%
    

\section{Experiments}
\subsection{Experimental Setting}
To evaluate the effectiveness of our approach, we conducte testing on six publicly available datasets. These include four synthetic distortion datasets: LIVE \coloredcite{1709988}, CSIQ \coloredcite{larson2010most}, KADID-10K \coloredcite{8743252} and TID2013 \coloredcite{PONOMARENKO201557} and three authentic distortion datasets including LIVEC \coloredcite{8478807}, SPAQ \coloredcite{9156490} and KonIQ \coloredcite{8968750}. Specifically, the LIVE dataset contains 779 distorted images covering 5 types of distortions; CSIQ includes 866 images with 6 distortion types; TID2013 comprises 3,000 images involving 24 distortion types; and KADID-10K consists of 10,125 images corresponding to 25 distortion types. In addition, LIVEC contains 1,162 images, KonIQ includes 10,073 images, and SPAQ comprises 11,125 images. We employ the Adam optimizer, setting the learning rate at $2\times10^{-4}$ for synthetic distortion datasets, and $2\times10^{-5}$ for authentic distortion datasets.Unless otherwise specified, the Transformer decoder comprises $4$ layers, the query set size is $6$, the MoE head uses $4$ experts with Top-$K{=}2$, and the decoder adopts multi-head attention with $6$ heads and an embedding dimension of $384$. The model is trained for 30 epochs; a batch size of 64 is used for small-scale datasets and 128 for large-scale datasets, and the learning rate follows a step schedule with a decay factor of $0.01$ applied every 10 epochs. Like other IQA methods, we evaluate our test results using Pearson linear correlation coefficient (PLCC) and Spearman rank order correlation coefficient (SROCC). Higher values for both metrics indicate better performance. Our method compared state-of-the-art algorithms including DBCNN \coloredcite{8576582}, MMMNet \coloredcite{9337868}, VCRNet \coloredcite{9694502}, MUSIQ \coloredcite{9376644}, TRes \coloredcite{9706735}, DACNN \coloredcite{9817377}, DEIQT \coloredcite{qin2023dataefficientimagequalityassessment}, CICI \coloredcite{10213444}, PCIQA \coloredcite{10520932}, TempQT \coloredcite{10287911}, FsPN \coloredcite{10706861}, LIQE \coloredcite{10204315}, Re-IQA \coloredcite{10204004}, QCN \coloredcite{10656546}, LoDa \coloredcite{10655501}, VISGA \coloredcite{10845143}. The input distorted images were randomly cropped into small segments measuring $224 \times 224$ pixels. We randomly split each dataset into $0.8$ training set and $0.2$ testing set, and repeated the experiments $10$ times, reporting the median values of SROCC and PLCC.

\begin{figure}[t]
\label{sec:intro}
\centering
\includegraphics[height=4cm]{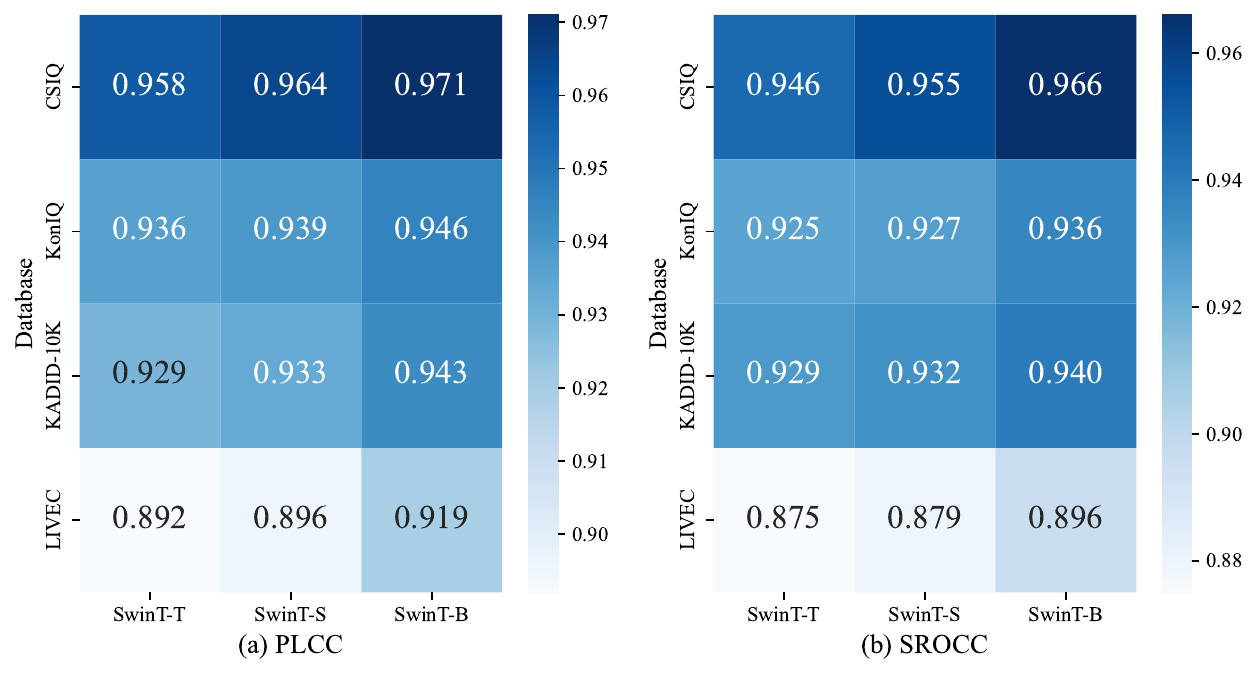}
\caption{Performance of different scale pretrained Swin Transformers across datasets (PLCC/SROCC).}
\vspace{-6mm}
\label{fig_6}
\end{figure}






\begin{figure*}[!t]
\label{sec:intro}
\centering
\includegraphics[width=7in]{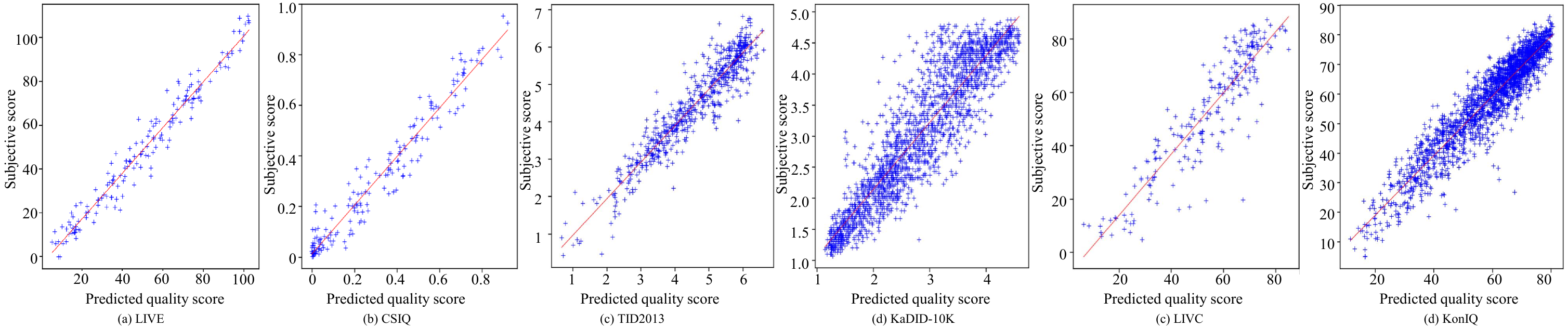}
\caption{Scatter plots of ground-truth against predicted scores of proposed Life-IQA on six datasets. }
\vspace{-4mm}
\label{fig_7}
\end{figure*}

\subsection{Comparisons with the State-of-the-art Methods}

Table~\ref{table_1} presents a comprehensive comparison between the proposed Life-IQA and several state-of-the-art IQA methods across seven benchmark datasets.
Life-IQA employs a Swin Transformer pretrained on ImageNet-21K as the backbone network and is further fine-tuned on ImageNet-1K, containing about 95M parameters. On CSIQ, Life-IQA achieves 0.966/0.971, surpassing DEIQT (24M) by 0.020/0.008 and slightly outperforming VISGA (0.960/0.971) with higher correlation consistency.
On TID2013, Life-IQA reaches 0.918/0.930, outperforming DEIQT (0.892/0.908) by 0.026/0.022 and VISGA (0.901/0.914) by 0.017/0.016.
On KADID-10K, Life-IQA attains 0.940/0.943, comparable to VISGA (0.939/0.944) and substantially higher than DEIQT (0.889/0.887).
On KonIQ, Life-IQA achieves 0.943/0.946, outperforming VISGA (0.931/0.940) by 0.012/0.006 and DEIQT (0.921/0.934) by 0.022/0.012.
On SPAQ, Life-IQA reaches 0.926/0.929, again exceeding DEIQT (0.919/0.923) and maintaining top-tier results.
Compared to compact architectures such as DEIQT (24M) and FsPN (34M), Life-IQA achieves notable accuracy gains.
Meanwhile, when compared with larger scale models like LIQE (151M) and TRes (152M), Life-IQA achieves comparable or superior performance while maintaining a more efficient parameter footprint.
Overall, Life-IQA delivers state-of-the-art performance across diverse datasets while maintaining acceptable computational complexity, showcasing a well-balanced design that combines expressive power with practical efficiency.

\subsection{Cross-Dataset Evaluation}
To validate the generalization capability of the Life-IQA model, we conduct cross-dataset validation by training on the KonIQ datasets and testing on several other datasets. As shown in Table \ref{table_3}, when trained on KonIQ, Life-IQA demonstrates strong cross-dataset generalization. Life-IQA attains 0.715/0.738 on CSIQ, 0.453/0.578 on TID2013, and 0.822/0.800 on LIVE. It ranks first on five of six metrics and remains competitive on the remaining one (SROCC on LIVE). The advantage appears in both SROCC and PLCC correlations, indicating reliable ordering and calibration across datasets with differing content and distortion regimes. Overall, these results highlight the strong cross-dataset generalization of Life-IQA.

\begin{figure}[t]
\label{sec:intro}
\centering
\includegraphics[height=5cm]{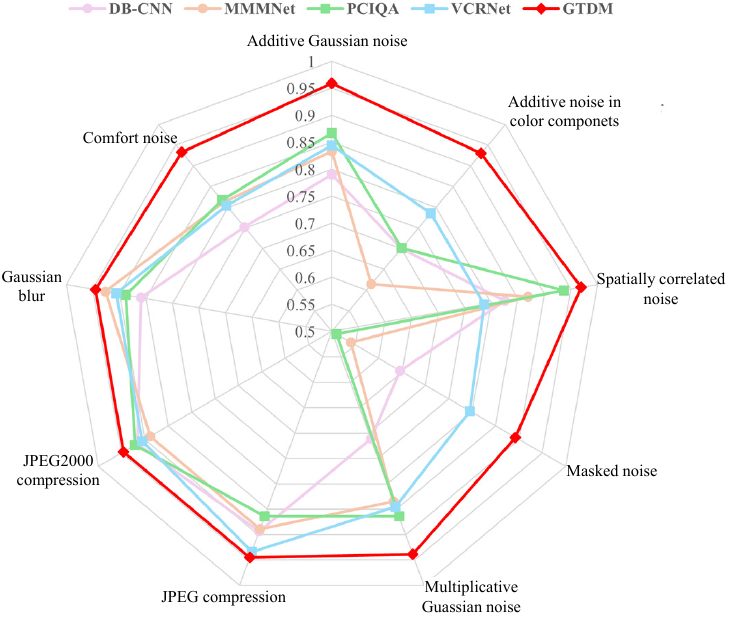}
\caption{Comparative analysis of the SROCC across various methods on the distortion types within TID2013.}
\vspace{-5mm}
\label{fig_8}
\end{figure}

\begin{table}[htbp]
  \centering
  \caption{Data-efficient learning validation with the training set
containing 20\%, 40\% and 60\% images.}
  \scalebox{0.85}{
  \renewcommand{\arraystretch}{1.2}
    \begin{tabular}{cp{7em}cccccc}
    \toprule
    \multicolumn{2}{c}{} & \multicolumn{2}{c}{KonIQ} & \multicolumn{2}{c}{LIVEC} & \multicolumn{2}{c}{LIVE} \\
\cmidrule{3-8}    \multicolumn{1}{p{1.5em}}{Mode} & \multicolumn{1}{c}{Methods} & \multicolumn{1}{p{2.5em}}{SROCC} & \multicolumn{1}{p{2.5em}}{PLCC} & \multicolumn{1}{p{2.5em}}{SROCC} & \multicolumn{1}{p{2.5em}}{PLCC} & \multicolumn{1}{p{2.5em}}{SROCC} & \multicolumn{1}{p{2.5em}}{PLCC} \\
    \midrule
    \multirow{3}[2]{*}{20\%} & HyperIQA \coloredcite{9156687} & 0.869 & 0.873 & 0.776 & 0.809 & 0.951 & 0.950 \\
          &  DEIQT \coloredcite{qin2023dataefficientimagequalityassessment} & 0.888 & 0.908 & 0.792 & 0.822 & 0.965 & 0.968 \\
          & Life-IQA  &  \textbf{\textcolor{red}{0.908}}     & \textbf{\textcolor{red}{0.923}}      & \textbf{\textcolor{red}{0.804}} & \textbf{\textcolor{red}{0.842}} &  \textbf{\textcolor{red}{0.970}}     & \textbf{\textcolor{red}{0.970}} \\
    \midrule
    \multirow{3}[2]{*}{40\%} & HyperIQA \coloredcite{9156687} & 0.892 & 0.908 & 0.832 & 0.849 & 0.959 & 0.961 \\
          &  DEIQT \coloredcite{qin2023dataefficientimagequalityassessment} & 0.903 & 0.922 & 0.838 & 0.855 & \textbf{\textcolor{red}{0.971}} & \textbf{\textcolor{red}{0.973}} \\
          & Life-IQA  &   \textbf{\textcolor{red}{0.920}}    &   \textbf{\textcolor{red}{0.930}}    &   \textbf{\textcolor{red}{0.855}}    &   \textbf{\textcolor{red}{0.880}}    &  \textbf{\textcolor{red}{0.971}}      &  0.972 \\
    \midrule
    \multirow{3}[2]{*}{60\%} & HyperIQA \coloredcite{9156687} & 0.901 & 0.914 & 0.843 & 0.862 & 0.960  & 0.960 \\
          &  DEIQT \coloredcite{qin2023dataefficientimagequalityassessment} & 0.914 & 0.931 & 0.848 & 0.877 & 0.972 & \textbf{\textcolor{red}{0.974}} \\
          & Life-IQA  &   \textbf{\textcolor{red}{0.930}}    &  \textbf{\textcolor{red}{0.939}}     &\textbf{\textcolor{red}{0.882}}      &\textbf{\textcolor{red}{0.898 }}      &  \textbf{\textcolor{red}{0.973}}      & \textbf{\textcolor{red}{0.974}}  \\
    \bottomrule
    \end{tabular}%
    }
  \label{table_4}
  \vspace{-5mm}  
\end{table}%

\subsection{Effectiveness of Life-IQA}

To evaluate the data efficiency of the proposed model, we conduct experiments on the LIVE, LIVEC, and KonIQ datasets by training with $20\%$, $40\%$, and $60\%$ of the original training data, respectively, and testing on the full test set. As shown in Table \ref{table_4}, Life-IQA achieves superior performance to DEIQT even with only $20\%$ of the training data on the synthetic distortion dataset LIVE. On the real-distortion datasets LIVEC and KonIQ, Life-IQA further surpasses DEIQT, demonstrating its strong learning capability under limited data conditions.
In addition, to examine the effect of model scale, Life-IQA is evaluated on Swin-Tiny, Swin-Small, and Swin-Base backbones, all pretrained on ImageNet-21K. As illustrated in Figure \ref{fig_6}, the performance of Life-IQA consistently improves as the backbone size increases, further validating the effectiveness of the proposed framework.
Figure \ref{fig_7} presents the scatter plots of subjective quality scores versus Life-IQA predictions, where the red line denotes the fitted regression curve. Across the LIVE, CSIQ, TID2013, LIVEC, and KonIQ datasets, Life-IQA exhibits a strong linear correlation between predicted and ground truth scores, with minimal dispersion, indicating its robust and accurate prediction capability for both synthetic and real distortions.
Furthermore, to evaluate the performance of Life-IQA on individual distortion types, we conduct experiments on nine representative distortion categories from the TID2013 dataset. As shown in the Figure \ref{fig_8}, Life-IQA significantly outperforms existing state-of-the-art methods on several distortion types, including additive gaussian noise, comfort noise, and masked noise, confirming its superior effectiveness in handling single distortion scenarios.

\begin{figure}[!t]
\label{sec:intro}
\centering
\includegraphics[height=7cm]{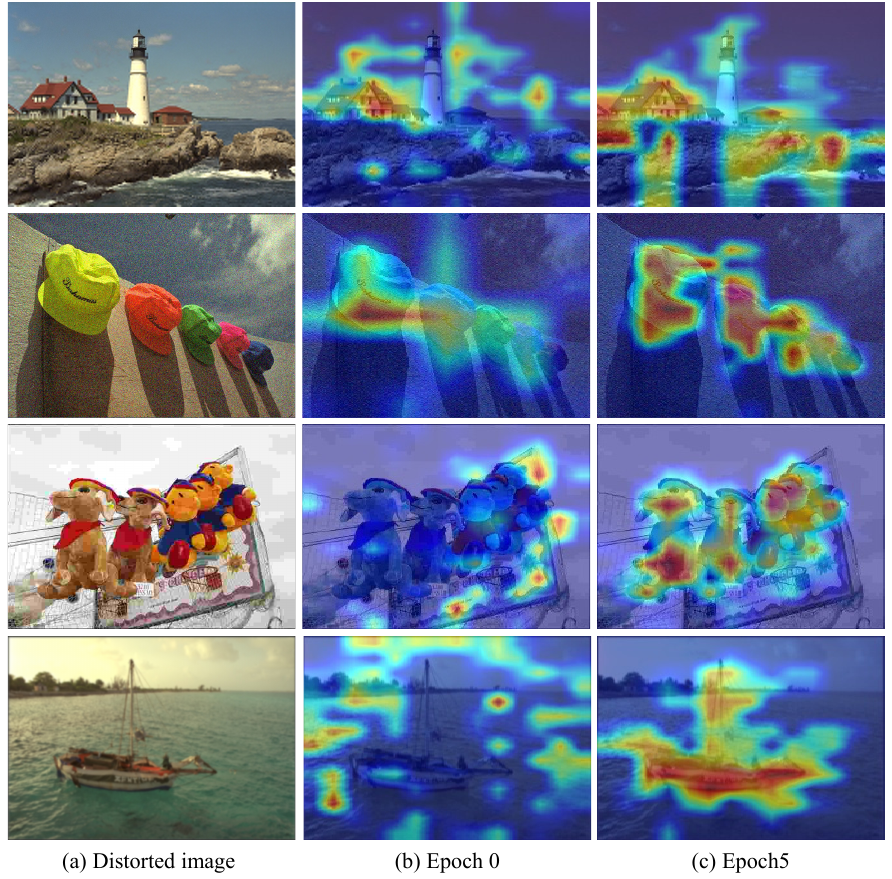}
\caption{Visualization results at different epochs for four test images with
different scene contents. }
\vspace{-5mm}
\label{fig_9}
\end{figure}

\begin{table}[htbp]
  \centering
  \caption{Decoder variant comparison. }
  \renewcommand{\arraystretch}{1}
  \scalebox{1}{
   \renewcommand{\arraystretch}{1}
    \begin{tabular}{p{6em}ccccc}
    \toprule
    \multirow{2}[4]{*}{Method} & \multirow{2}[4]{*}{Param} & \multicolumn{2}{c}{CSIQ} & \multicolumn{2}{c}{KADID-10K} \\
\cmidrule{3-6}
        &        & \multicolumn{1}{p{2em}}{SROCC} & \multicolumn{1}{p{2em}}{PLCC} & \multicolumn{1}{p{2em}}{SROCC} & \multicolumn{1}{p{2em}}{PLCC} \\
    \midrule
    GCN $\rightarrow$ MHA  & 96M & 0.942 & 0.954 & 0.917 & 0.914 \\
    MoE $\rightarrow$ FFN  & 99M & 0.952 & 0.960 & 0.924 & 0.925 \\
    SwinT+decoder          & 96M & 0.959 & 0.968 & 0.926 & 0.926 \\
    \midrule
    Life-IQA               & 95M & \textbf{0.966} & \textbf{0.971} & \textbf{0.940} & \textbf{0.943} \\
    \bottomrule
    \end{tabular}%
  }
  \label{table_5}
  \vspace{-4mm}
\end{table}

\begin{table}[!t]
  \centering
\caption{Cross-layer pairing study: deep and shallow interactions}
  \renewcommand{\arraystretch}{1}
    \begin{tabular}{p{6.6em}cccc}
    \toprule
    \multirow{2}[4]{*}{Method} & \multicolumn{2}{c}{CSIQ} & \multicolumn{2}{c}{KADID-10K} \\
\cmidrule{2-5}    \multicolumn{1}{c}{} & \multicolumn{1}{c}{SROCC} & \multicolumn{1}{c}{PLCC} & \multicolumn{1}{c}{SROCC} & \multicolumn{1}{c}{PLCC} \\
    \midrule
    Stage2$\Longleftrightarrow $Stage1 &  0.951     &  0.959     & 0.915      & 0.918 \\

    Stage3$\Longleftrightarrow $Stage1 &0.960       &0.965       & 0.937      &0.940  \\
    
    Stage3$\Longleftrightarrow $Stage2 & 0.962      &  0.968     &  0.938     & 0.938 \\
   
    Stage4$\Longleftrightarrow $Stage1 &  0.958     &   0.965    & 0.930      &0.930  \\

    Stage4$\Longleftrightarrow $Stage2 &  0.964     &   0.968    &  0.937     & 0.939 \\
    
    Life-IQA &  \textbf{0.966}     &   \textbf{0.971}    &   \textbf{0.940}    & \textbf{0.943} \\
    \bottomrule
    \end{tabular}%
     \vspace{-4mm}
  \label{table6}%
\end{table}%


\subsection{Visual Analysis}
Figure~\ref{fig_9} visualizes the attention maps of Life-IQA at different training epochs for four test images with diverse scene contents. At the Epoch 0, the attention responses are relatively scattered and fail to capture the essential structures of the scene.
The model tends to focus on background regions or exhibit random activations, indicating that the feature representations are still under initialization and have not yet learned the perceptual cues related to image quality.
As the Epoch 5, the attention maps become more compact and semantically meaningful.
Life-IQA effectively highlights the visually important regions, such as the main objects (e.g., the lighthouse, the hats, and the boat) and the distorted areas that strongly influence perceived quality.
This demonstrates that the Life-IQA gradually learns to align its focus with human perception, capturing both content-aware and distortion-sensitive information.


\begin{table}[htbp]
  \centering
   \renewcommand{\arraystretch}{1.2}
   \caption{Ablation Study Results on CSIQ and KADID-10K.}
  \scalebox{1}{
   \renewcommand{\arraystretch}{1}
    \begin{tabular}{p{6em}cccc}
    \toprule
    \multirow{2}[4]{*}{Method} & \multicolumn{2}{c}{CSIQ} & \multicolumn{2}{c}{KADID-10K} \\
\cmidrule{2-5}    \multicolumn{1}{c}{} & \multicolumn{1}{p{3em}}{SROCC} & \multicolumn{1}{p{3em}}{PLCC} & \multicolumn{1}{p{3em}}{SROCC} & \multicolumn{1}{p{3em}}{PLCC} \\
    \midrule
    {w/o stage3} & 0.938 & 0.952 &  0.933     & 0.934 \\
     {w/o MoE} & 0.934 & 0.949 & 0.920  & 0.921 \\
     {w/o GCN} & 0.952 & 0.963 & 0.931 & 0.931 \\
    \midrule
     {Life-IQA} & \textbf{0.966} & \textbf{0.971} & \textbf{0.940}  & \textbf{0.943} \\
    \bottomrule
    \end{tabular}%
    }
\vspace{-5mm}
  \label{table_6}%
\end{table}%
\subsection{Ablation Study}
Under a unified training protocol and matched parameter budgets, Table~\ref{table_5} isolates where the improvement comes from by constraining substitutions within the Life-IQA framework. The comparison includes: the vanilla SwinT+Transformer decoder; a Life-IQA variant in which the query topology GCN is replaced by the vanilla multi head attention while keeping the same decoder placement; a Life-IQA variant in which the head only MoE is replaced by a standard FFN; and the full Life-IQA. Both single substitutions yield consistent accuracy declines across CSIQ and KADID-10K, and the vanilla decoder remains inferior to the full model. These results indicate that the gains arise from the proposed decoder design and the specific placement of GCN and MoE, rather than from simple component swapping or additive combinations.
Table~\ref{table6} varies only the interaction pair in the GCN-enhanced layer interaction, while the decoder depth, heads, embedding size, query count, training recipe, and parameter budget are kept fixed. A consistent pattern emerges:
Using Stage4 with Stage3 (Life-IQA) achieves 0.966/0.971 on CSIQ and 0.940/0.943 on KADID-10K.
Relative to Stage4 with Stage1, the deep–deep pairing gains +0.008/+0.006 SROCC/PLCC on CSIQ and +0.010/+0.013 on KADID-10K.
 Compared with Stage4 with Stage2, the deep–deep pairing improves by +0.002/+0.003 on CSIQ and +0.003/+0.004 on KADID-10K; compared with Stage3 with Stage2, the gains are +0.004/+0.003 on CSIQ and +0.002/+0.005 on KADID-10K.
The Stage2$\Leftrightarrow$Stage1 pairing yields 0.951/0.959 on CSIQ and 0.915/0.918 on KADID-10K, underperforming the deep–deep setting by -0.015/-0.012 (CSIQ) and -0.025/-0.025 (KADID-10K).
The trend holds for both rank correlation (SROCC) and linear correlation (PLCC), indicating that interactions between semantically compatible deep representations yield more reliable quality cues, whereas shallow–shallow coupling provides limited semantic separation and thus the weakest performance.
Table~\ref{table_6} assesses component removal inside Life-IQA. Eliminating the MoE head, eliminating the query topology GCN, or removing the deep feature–guided fusion path (i.e., the $stage3$ K/V pathway in the deep decoder interaction) each produces a clear and reproducible degradation on both datasets. The evidence supports that the expert head, the query topology modeling, and the deep decoder interaction via $stage4$→query and $stage3$→key/value are all necessary contributors to the overall effectiveness of Life-IQA.

  
    
  


\section{Conclusion}
This paper investigates the role of deep feature fusion in image quality assessment and introduces the Life-IQA framework. The Life-IQA leverages the Transformer decoder to integrate multi-level features, employs GCN to enhance intra-query interaction and perception, and utilizes MoE for multi-dimensional feature modeling. Extensive experiments conducted on seven datasets demonstrate that Life-IQA achieves superior performance across various scenarios. 


\bibliographystyle{IEEEtran}
\bibliography{REFERENCES}

\vfill

\end{document}